\pdfoutput=1

\documentclass[11pt]{article}
\usepackage[table]{xcolor}

\usepackage[preprint]{acl}

\usepackage{times}
\usepackage{latexsym}
\usepackage{comment}

\usepackage[T1]{fontenc}

\usepackage[utf8]{inputenc}

\usepackage{microtype}

\usepackage{inconsolata}

\usepackage{graphicx}

\usepackage{amsmath}
\usepackage[noabbrev]{cleveref}
\usepackage{booktabs}
\usepackage{multirow}
\usepackage{xspace}
\usepackage{tabularx}
\usepackage{hyperref}
\usepackage{tcolorbox}
\usepackage{fancyhdr}
\usepackage{float}
\tcbset{
    rounded corners,
    colback = white,
}

\title{Estimating LLM Consistency: A User Baseline vs Surrogate Metrics}

\author{
  \textbf{Xiaoyuan Wu\textsuperscript{1}},
  \textbf{Weiran Lin\textsuperscript{1}},
  \textbf{Omer Akgul\textsuperscript{2, 1}},
  \textbf{Lujo Bauer\textsuperscript{1}}
  \\
  \textsuperscript{1}Carnegie Mellon University
  \textsuperscript{2}RSAC Labs
  \\
  \small{
    \textbf{Correspondence:} \href{mailto:wxyowen@cmu.edu}{wxyowen@cmu.edu}
  }
}

\begin{document}
\maketitle
\newcommand{\lmsys}{\emph{LMSYS}\xspace}
\newcommand{\coqa}{\emph{CoQA}\xspace}

\newcommand{\gemma}{Gemma\xspace}
\newcommand{\llama}{Llama\xspace}
\newcommand{\mistral}{Mistral\xspace}

\newcommand{\bleu}{\emph{BLEU}\xspace}
\newcommand{\bert}{\emph{Bert}\xspace}
\newcommand{\rouge}{\emph{Rouge}\xspace}
\newcommand{\usescore}{\emph{USE}\xspace}
\newcommand{\semanticentropy}{\emph{SE}\xspace}

\newtcolorbox{resultbox}{
    boxrule = 0.5pt,
    colframe = black,
    top = 0pt,
    bottom = 0pt,
    left = 2pt,
    right = 2pt
}

\begin{abstract}
    Large language models (LLMs) are prone to hallucinations and sensitive
    to prompt perturbations, often resulting in inconsistent or unreliable
    generated text. Different methods have been proposed to mitigate such
    hallucinations and fragility, one of which is to measure the
    consistency of LLM responses---the model's confidence in the response
    or likelihood of generating a similar response when resampled. In
    previous work, measuring LLM response consistency often relied on
    calculating the probability of a response appearing within a pool of resampled
    responses, analyzing internal states, or evaluating logits of resopnses.
    However, it was not clear how well these
    approaches approximated users' perceptions of consistency of LLM
    responses. To find out, we performed a user study ($n=2,976$)
    demonstrating that current methods for measuring LLM response
    consistency typically do not align well with humans' perceptions of LLM
    consistency. We propose a logit-based ensemble method for estimating
    LLM consistency and show that our method matches the performance of the
    best-performing existing metric in estimating human ratings of LLM
    consistency. Our results suggest that methods for estimating LLM
    consistency without human evaluation are sufficiently imperfect to
    warrant broader use of evaluation with human input; this would avoid
    misjudging the adequacy of models because of the imperfections of
    automated consistency metrics.
\end{abstract}

\section{Introduction}\label{sec:intro}

Large language models (LLMs)
have seen rapid adoption across a multitude of domains despite numerous
inherent limitations, such as hallucinating responses and
being fragile to adversarial inputs. 
Hallucinations can have disastrous consequences in high-stakes fields
like healthcare and law~\cite{merken2025hallucinations}, 
prompting concern even as 
adoption continues~\cite{bbc2025hilldickinson, metnick2024aihealthcare}.
Further, the fragility of LLMs enables a range of 
misuses %
that could harm users~\cite{kumar2024manipulating, lin_llm_2025}. 
For instance, \citet{lin_llm_2025} showed that minor, unnoticed changes to suggested prompts 
can result in outputs with biases 
controlled by an adversary.

Researchers have suggested that some of these issues correlate with the
inconsistency of LLMs, which is generally defined as their tendency to
generate low-confidence responses or conflicting responses when the same
prompt is resampled~\cite{manakul_selfcheckgpt_2023}.
Accurately estimating LLM consistency is important because it can support
multiple critical applications: predicting whether an answer is factual,
thereby improving reliability in high-stakes
domains~\cite{duan_shifting_2024}; detecting fragile or malicious
prompts~\cite{lin_llm_2025}; informing membership-inference
attacks~\cite{mattern_etal_2023_membership}; or signaling to users the
level of trust they should place in LLM
outputs~\cite{kapoor_large_2024,Ruggieri_Pugnana_2025}.

Consequently, many attempts have been made
to define and measure the consistency of LLMs. This body of work 
can roughly be divided into two categories: (1) estimating consistency based 
on LLMs' internal states or logits and (2) estimating consistency 
based on resampling LLMs' responses. While the former is more computationally efficient, 
the latter is empirically well-grounded~\cite{kuhn_semantic_2023, qiu_semantic_2024} and applicable even without white-box access to a model~\cite{lin_generating_2024}.

Sampling-based estimation methods fundamentally rely on sampling from an LLM--prompt
pair and comparing the outputs using a comparison function. This function differs between 
works---e.g.,~\citet{duan_shifting_2024} 
vs~\citet{manakul_selfcheckgpt_2023}---but partially boils down to 
estimating semantic similarity of responses. 
However, to our knowledge,
none have based their estimation of LLM model consistency 
on user-based comparisons, the ground truth for 
semantic similarity~\cite{bowman2015large, agirre_semeval-2014_2014}.
Further, existing uncertainty estimation methods have not yet
 demonstrated reliability or alignment with human judgments of
 LLMs' consistency.
Without a baseline,
consistency metrics are typically evaluated by their ability to
predict whether a
model output is factual~\cite{kuhn_semantic_2023, qiu_semantic_2024, duan_shifting_2024, zhang_luq_2024, kapoor_large_2024}.
Joining recent work~\cite{novikova2025consistency}, 
we argue that consistency might be able to predict accuracy but is fundamentally an 
independent property of the model-prompt pair.
There should be a metric that conveys 
to what degree a model, when provided a prompt, will repeatedly produce responses with equivalent meaning.
We further posit that the baseline for measuring consistency is reliant on the
comparison of semantics, the ground truth of which is defined by human
judgement~\cite{bowman2015large,
agirre_semeval-2014_2014,nguyen_using_2014,raj_semantic_2025}.

We aim to fill this gap by using user-based comparisons to estimate 
the consistency of LLMs, and by investigating whether an automated metric can come close to simulating users' judgments. 
Specifically, we conduct a user study with 2,976 participants to collect 
semantic similarity ratings between 
a sample of 10 responses to each of 100 prompts, totaling 14,880 comparisons.
We then calculate response-level (one score per response) and 
prompt-level (one score per prompt) consistency, establishing a user-based
LLM consistency baseline.
Through a series of experiments, we show that existing metrics of measuring
LLM response uncertainty do not align well with human judgements collected
in this study. We further show that an ensemble of logit-based scores is
as similar to human judgement 
as the best-performing of the other methods we tested.
We find that the discrepancy between existing metrics and human judgements fluctuates 
between models and between datasets, with previous 
metrics being less similar to human judgements
on real-world prompts then on synthetic prompts. Based on our results, we
advocate for more human-based LLM response consistency evaluation in future work.

We structure the remainder of the paper in four sections:
\cref{sec:rel-work} details background work; \cref{sec:methods} describes
our model- and prompt-selection strategy, user study, and consistency
calculation; \cref{sec:results} reports our experiments and comparisons;
finally, we summarize and discuss implications of our work in
\cref{sec:discussion}.

\begin{figure*}[t]
      \centering
      \includegraphics[width=\linewidth]{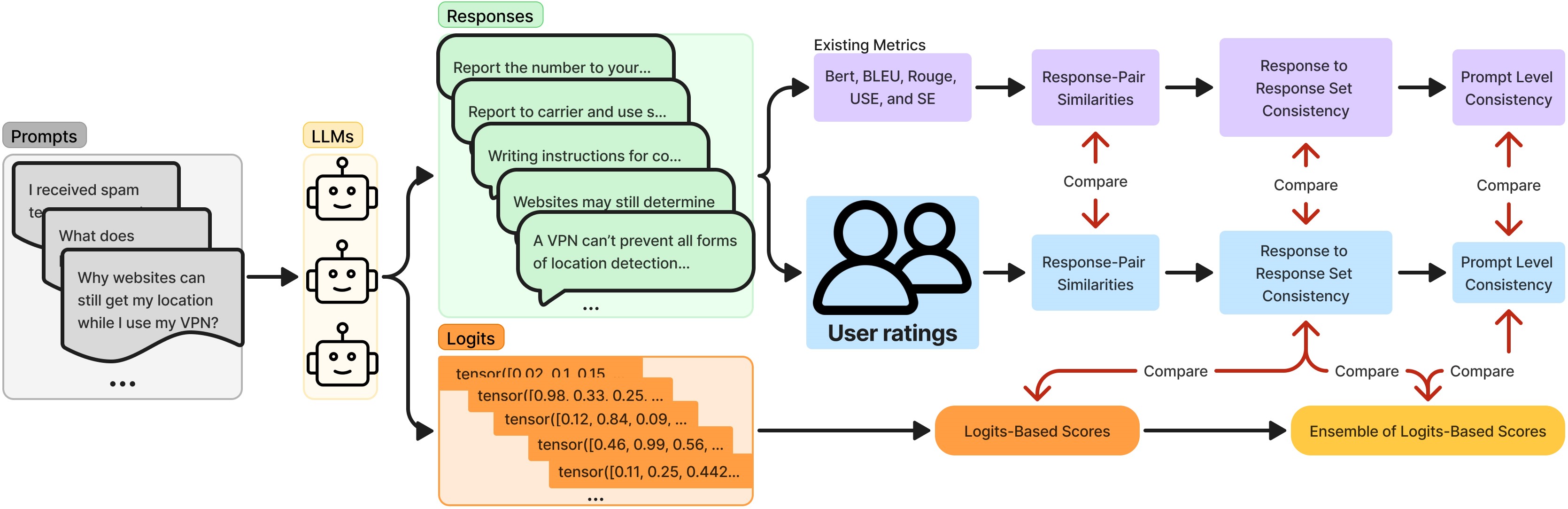}
      \caption{Overview of our study design}
      \label{fig:study-overview}
\end{figure*}

\section{Related Work}\label{sec:rel-work} Here, we
review work in consistency estimation of LLMs with resampling and internal model states.
(\ref{subsec:rel-work:llm-consistency}).
We also cover work that investigated
logit-based metrics to estimate LLM consistency
(\ref{subsec:rel-work:machine-metrics}).
Finally, we reiterate our study motivation (\ref{subsec:rel-work:user-important}).

\subsection{LLM Consistency from Sampling and Internal States }\label{subsec:rel-work:llm-consistency}

Prior definitions of model consistency \cite{lakshminarayanan2017simple} do not apply to the 
near-infinite output space, auto-regressive nature of LLMs~\cite{kuhn_semantic_2023}. 
As a result, a line of work has emerged to define uncertainty of LLM responses 
(sometimes defined as the confidence of the model in a response, 
or how likely a response is to be generated when resampled). 

\paragraph{Sampling} When estimating uncertainty, 
researchers often resample multiple responses from a model-prompt 
pair. A chosen response is then compared to the set of responses, ultimately calculating an uncertainty score. 
For instance,~\cite{kuhn_semantic_2023} 
establishes a set of semantically equivalent 
responses within the sampled set using entailment. 
The probabilities of any one of these semantically 
equivalent groups are fed 
into a predictive entropy based formulation 
to obtain an uncertainty score. Follow-up work has 
suggested similar sampling-based metrics~\cite{qiu_semantic_2024, duan_shifting_2024}.

\paragraph{Internal state representation} Some prior work has sidestepped 
the definition of uncertainty to directly predict if a given answer is 
right or wrong. This body of work (sometimes also referred to as uncertainty estimation), 
uses techniques such as asking follow-ups~\cite{kadavath_language_2022, sam2025predicting} and 
training auxiliary prediction models based on internal states~\cite{kapoor_large_2024, kossen2024semantic}.

\subsection{Probability- and Logit-Based Uncertainty
Metrics}\label{subsec:rel-work:machine-metrics} Existing work suggests that
the
confidence of individual token generation can suggest the consistency of LLM
responses~\cite{manakul_selfcheckgpt_2023}. Borrowing 
definitions from~\citet{manakul_selfcheckgpt_2023}
, we denote the probability
distribution of an individual token generated as $p$, and the entropy of
individual token generation $H$, $H=-\sum_i p_i log(p_i)$. Existing metrics
include the average of the minus log probability $Avg(-log(p))$, maximum of the minus log
probability $Max(-log(p))$, average of entropy $Avg(H)$ and maximum of entropy
$Max(H)$. In this paper, we ensemble these metrics (and variants) in
\ref{subsec:methods:machine-metrics}.

\paragraph{Difference of Logits Ratio Loss} \label{subsec:rel-work:DLR} Besides
probability-based (i.e., $p$-based) uncertainty metrics, we also find
logit-based uncertainty metrics to suggest LLM response consistency. We denote
the logits of individual token generation as $l$, $p=softmax(l)$. Specifically,
we denote the largest, second largest, third largest, and fourth largest logit
of a token generation as $l_1$, $l_2$, $l_3$,  and $l_4$ correspondingly. The
Difference of Logits Ratio Loss (DLR loss), formally $\frac{l_1 -
l_2}{[l_1-(l_3+l_4)]/2}$, is a loss function commonly used by adversarial
attacks to suggest how confident a model is~\cite{croce_reliable_2020}. 
In this paper, we borrow the DLR loss as a
logit-based uncertainty of a token generation to estimate the consistency of LLM
responses. 

Since probability-based uncertainty estimation is ultimately derived from 
token logits, for the rest of the paper, we call this group of uncertainty estimation 
logit-based methods.

\subsection{Importance of User Evaluation in NLP
Tasks}\label{subsec:rel-work:user-important} Human evaluations are important in
improving and assessing the quality of 
natural language processing (NLP)~\cite{gritta_humanrankeval_2024,
boyd-graber_human-centered_2022,blodgett_human-centered_2024}. 
Researchers have
investigated how well machines are able to translate sentences into another
language, compare sentences for semantic similarity, and many other NLP tasks
with human evaluations~\cite{chatzikoumi_how_2020, graham_can_2017}. 
This trend has continued in the LLM era~\cite{ouyang2022training, bai2022training}
While prior work has found ways to estimate the consistency
of responses produced by LLMs,
we argue that consistency is defined by what 
degree humans see responses as similar to each other. Using this definition, 
we establish a baseline through a user study.

\section{Methods}\label{sec:methods} 
In this study, we seek to understand how accurately existing metrics can approximate
users' perceptions of LLM consistency. We explore ways to create novel and more
efficient ways of estimating LLM consistency without the need for expensive user
studies or generating multiple responses per prompt. We provide an overview of
our study design in~\cref{fig:study-overview}.

In this section, we first describe the prompts, models, and responses used
in the study (\ref{subsec:methods:prompts-responses}). We then describe in
detail how we conducted the user study with 2,976 participants
(\ref{subsec:methods:user-study}).
Finally, we explain how we calculate consistency, both with existing metrics and 
our user-based method (\ref{subsec:methods:machine-metrics}).

\subsection{Prompts, Models, and
Responses}\label{subsec:methods:prompts-responses}

\paragraph{Prompts}
Following prior work~\cite{geng_survey_2024}, we selected two prompt sets: (1) to ensure comparability with prior
research, we use a dataset of open-ended questions from
\coqa~\cite{reddy_coqa_2019}, which is commonly adopted in the literature;
and (2) to better represent real-world use cases, we include a sample of 50
prompts from LMSYS-Chat-1M~\cite{zheng_lmsys-chat-1m_2023}.

Each entry in the \coqa dataset contains a story with a series of questions
about the story. For our study, we only used the first question in each
series, as many following questions are dependent on the previous question
or answer for context. We randomly sampled 50 story--question pairs from the
\coqa dataset.

Real-world prompts from the \lmsys dataset are more diverse than
those in \coqa. Hence, we used labels in the data to filter out prompts
that are were not in English or \emph{flagged}\footnote{by OpenAI
Moderation~\cite{zheng_lmsys-chat-1m_2023}} (e.g., harassment, violence).
Next, we randomly sampled 100 prompts from the filtered \lmsys dataset and
manually removed 48 prompts: 14 coding questions, 9 about the LLM itself, 9 that were inappropriate (e.g.,
sexual), 6 that contain nonsensical sentences, 5 that we
expected to have non-English responses (e.g., translations), 3
asking for answers over 1,500 words, and 2 that contain
time-sensitive information. We excluded coding 
questions since participants might not have the background to
rate coding-related responses. We
randomly removed 2 prompts from the remaining 52 to obtain a final sample
of 50.

\paragraph{Models} 
A generalizable definition of consistency should be model-agnostic. Thus, we
used three open-weights models from competing institutions to generate
responses: Llama-3.2-3B-Instruct (\llama), Gemma-2-9B-it (\gemma), and
Mistral-7B-Instruct-v0.3 (\mistral). Each was the most recent publicly
available version of their repective families, was widely used in previous
work, and met our hardware memory constraints (Nvidia RTX A6000). 
We used open-weights models as they provided us access to logits, which were
necessary for our calculation described
in~\cref{subsec:methods:machine-metrics}, and enabled reproducibility of
our study~\cite{ma2024schrodinger}. 
We left temperature settings unchanged from the
default configuration because there is no consensus from pervious
work on optimal temperature for uncertainty
estimation~\cite{cecere_monte_2025,du_optimizing_2025,wang_self-consistency_2023,renze_effect_2024,zou_universal_2023}

\paragraph{Responses} The 100 prompts were randomly assigned to the three
LLMs; \gemma received 34, \llama 33, and \mistral 33.  Using the assigned
model, we then generated 10 responses per prompt, totaling 1,000 responses.
We took this approach as previous work showed that a set of 10 responses adequately
captures the semantic diversity of the response space for uncertainty
estimation purposes~\cite{kuhn_semantic_2023,qiu_semantic_2024}. Response 
generation took 30 minutes of GPU time.

\subsection{User Study}\label{subsec:methods:user-study}
While many metrics have been created to approximate LLM
consistency, a reporting of LLM consistency based on users' perceptions
has been largely missing
from the literature.
Our user-rating based approach establishes a
baseline against which other metrics can be evaluated.
We recruited users from Prolific and required that
participants be 18 years or older, reside in the United States, read and type
in English fluently, and have at least a 95\% approval rate on the platform.
We use Prolific over MTurk because recent work has shown its superior 
data quality~\cite{tang_replication_2022,phd_after_2018}.

Each participant was given an introduction to the study before seeing 
instructions for, and examples of, how to rate semantic similarity between a pair of
sentences using a 6-point scale (Appendix \ref{subsec:appendix:user-study-inst}). We adopted this
commonly-used scale,
with explanations and examples, 
from~\citet{agirre_semeval-2014_2014}.
After the instructions, each participant rated the semantic similarity of
five pairs of responses
from five different prompts. Participants were compensated \$1; the median
survey completion time was 5 minutes and 13 seconds. We collected participants'
Prolific IDs for compensating those who completed our survey. We
removed Prolific IDs before doing any analysis.
Lastly, we collected participants' demographics. The sentence pairs and the
order in which they were shown to participants were randomized. Our study was approved
by our institution's ethics review board.

\begin{table}
    \centering
    \footnotesize
    \begin{tabular}{ccccc}
        \toprule
        \textbf{Model} & \textbf{Num.}    & \textbf{Num.}        & \textbf{Num. Pairs of}   \\
        \textbf{Name}  & \textbf{Prompts} & \textbf{Responses}   & \textbf{Unique Responses} \\
        \midrule
        \textbf{\gemma}          & 34               & 340                  & 793                   \\
        \textbf{\llama}          & 33               & 330                  & 826                   \\
        \textbf{\mistral}        & 33               & 330                  & 1,319                  \\
        \bottomrule
    \end{tabular}
    \caption{Breakdown of number of prompts, responses, and pairs of unique responses per model.}
    \label{tab:models-prompts-responses}
\end{table}

For each prompt, we generated 10 responses, yielding
$\binom{10}{2} = 45$ pairs of responses.
With 100 prompts, we obtained a total
of 4,500 pairs of responses, of which
2,938 were unique. More details about the number of prompts, responses, and
pairs of unique responses for each model are described
in~\cref{tab:models-prompts-responses}.

\subsection{Calculating and Comparing Consistency}
\label{subsec:methods:machine-metrics}
Here, we outline how we calculated consistency scores
(we calculated similarities between pairs of responses, consistency for each
response, and consistency for each prompt) and compared user-based scores
to previous work.
\paragraph{User ratings of similarities} We first aggregate user ratings of
similarity between a pair of responses $r_a$ and $r_b$ by averaging
ratings of $n\geq 5$ participants after removing the highest and lowest scores~\cite{curran_methods_2016}.
Formally,
\begin{equation}\label{equation:sim-user}
    s(r_a, r_b)=\frac{\sum\limits_{k=1}^{n-2}h_k}{n-2}
\end{equation}

\paragraph{Response-pair level comparison} To understand how existing metrics
used for LLM consistency compare to human ratings, we compared them at the
response-pair level. For each prompt (e.g., ``What is Wh in batteries?''), 10
responses yielded 45 response-pairs, totaling 4,500 response-pairs for
the 100 prompts in our study.
For these response-pairs, we calculated the
Spearman correlation coefficient $\rho$ between user ratings of similarity
($s_{\text{user}}$)
and each of
the four existing metrics: \bert ($s_{\text{\bert}}$), \bleu ($s_{\text{\bleu}}$),
\rouge ($s_{\text{\rouge}}$), and \usescore ($s_{\text{\usescore}}$).
We used the Spearman correlation coefficient $\rho$ because
human ratings were on a 6-point Likert scale, and we did not assume linear correlation between human
ratings and existing metrics. We share our findings on how each of these
compare to human ratings in~\cref{subsec:results:sentence-pair}.

We calculated $s_{\text{\bert}}$
using the \emph{BERTScore} python
package\footnote{\url{https://pypi.org/project/bert-score/0.3.0/}},
$s_{\text{\bleu}}$ using the \emph{NLTK} python
package\footnote{\url{https://www.nltk.org/_modules/nltk/translate/bleu_score.html}},
$s_{\text{\rouge}}$ using the \emph{rouge-score} python
package\footnote{\url{https://pypi.org/project/rouge-score/}}, and
$s_{\text{\usescore}}$ using the \emph{universal-sentence-encoder} model on
Kaggle\footnote{\url{https://www.kaggle.com/models/google/universal-sentence-encoder}}.

\paragraph{Response to response-set consistency}
We used the similarity scores between pairs of
responses to calculate the consistency between a specific response $r_i$ and all the remaining $m-1$ responses. We call this \emph{response to response-set} consistency, and compute it by averaging the response-to-response similarities between the specific response and each of the other responses.
Formally,
\begin{equation}\label{equation:response-to-response-set}
    C(\text{prompt}, r_i) = \frac{\sum\limits_{\substack{j=1\\j\neq i}}^{m}s(r_j, r_i)}{m-1}
\end{equation}
In addition to the four metrics used in response-pair level comparison, we
add Semantic Entropy (\semanticentropy)~\cite{kuhn_semantic_2023} at this
level of comparison, as their definition of consistency is measured at the
response to response-set level.
To understand how well these five metrics perform at this level, we compute the
Spearman correlation coefficient 
and the mean squared error between each of the metrics and human ratings
in~\cref{subsec:results:response-consistency}. 

Additionally, motivated by reducing the overhead of resampling, we compare
logits-based scores to human ratings and propose an ensembling method that uses
a linear combination of logits
in~\cref{subsec:results:response-consistency}.

To calculate \semanticentropy, we used the \emph{roberta-large-mnli} model to
check entailment between pairs of
responses.\footnote{\url{https://huggingface.co/FacebookAI/roberta-large-mnli}}
We provide more details on logit-based scores and our ensembling method
in~\autoref{para:methods:logits}.

\paragraph{Prompt level consistency} 
Building on response to response-set consistency, we define \emph{prompt
consistency} as the average of the $m$
response to response-set consistency values of
all responses to the prompt; formally,
\begin{equation}\label{equation:prompt-consistency}
    C(\text{prompt}) = \frac{\sum\limits_{i=1}^{m}C(\text{prompt}, r_i)}{m}
\end{equation}
We use the Spearman correlation and mean squared error to determine how well
each of the existing metrics and our ensemble approximate human
ratings of prompt level consistency. We share our findings
in~\cref{subsec:res:consistency-by-prompt}.

\paragraph{Estimating consistency with logit-based
metrics}\label{para:methods:logits}
We estimate the consistency of LLM responses using token-based
uncertainty metrics used in previous work~\cite{manakul_selfcheckgpt_2023}.
Specifically, we use four uncertainty metrics: the probability, minus log
probability, entropy, and the DLR loss (introduced in
\ref{subsec:rel-work:DLR}). For each response, we measure the maximum, sum,
minimum, and average of these four token-based metrics, totaling $16$ values.
The maximum and minimum show the ranges on individual tokens, while the
average is the expected value across all tokens.
The sum suggests a cumulative estimate over the whole response, 
as different responses may have different lengths.
Unlike existing work that detected hallucination based producing a single
response~\cite{manakul_selfcheckgpt_2023}, we collect several responses to
estimate LLM consistency for a prompt, corresponding to the
definition in~\cref{equation:prompt-consistency}.

\paragraph{Ensembling uncertainty metrics}\label{para:methods:ensemble} In
addition to individual uncertainty-based metrics described in the previous
paragraph, we create an ensemble of these in an attemp to better approximate human ratings
of LLMs' consistency.
We used \emph{Sequential Feature
Selection}~\cite{rucksties_sequential_2011}
to determine the most important metrics before
composing them into an ensemble.

\section{Results}\label{sec:results}

Here, we first summarize the user-study data collection
(\ref{subsec:results:user-overview}), the results of which were used to
establish our baseline for consistency. To compare previous consistency
methods to our user-rating-based method, we followed a bottom-up approach.
First, we looked at how methods for estimating similarity between sentences
(or LLM responses, in our case) compare to user ratings
(\ref{subsec:results:sentence-pair}). Next, we used these similarity scores
(including user-based scores) to compute consistency scores for individual
responses (\ref{subsec:results:response-consistency}). We also compared
logit-based uncertainty-estimation methods to our user baseline. Finally,
we aggregated the consistency scores within a set of responses to obtain a
prompt-level consistency score (\ref{subsec:res:consistency-by-prompt}).
For each step of the hierarchy, we compared automated methods for
estimating consistency to our user-rating-based baseline, which we believe
is the closest to representing ground truth. We found that while some
methods outperform others, none are very close to the
user-rating-based baseline. We further found that our ensemble logit-based
scores approximates the best-performing sampling method with the benefit of
not needing pools of resampled responses.

\subsection{User Study Overview and
Demographics}\label{subsec:results:user-overview} We recruited 2,976
participants from Prolific, asking each to rate five unique response pairs
from different prompts. About 52\% of participants were female and 47\%
male. Participants' ages ranged from 18 to 88 with various education, ethnicity,
and income backgrounds. More details are provided in~\cref{tab:demo}.

\begin{table}
    \centering
    \footnotesize
    \begin{tabular}{l|cc|cc}
    \toprule
    Model   & \multicolumn{2}{c}{\textbf{\coqa}} & \multicolumn{2}{c}{\textbf{\lmsys}} \\
    & Prompt & Reponse & Prompt & Reponse \\
    \midrule
    \textbf{\gemma}   & 14.00 & 18.96            & 11.00 & 37.08             \\
    \textbf{\llama}   & 14.44 & 18.54            & 11.59 & 46.29            \\
    \textbf{\mistral} & 14.69 & 49.18            & 13.85 & 133.83            \\
    \bottomrule
    \end{tabular}
    \caption{Average prompt and response lengths for each model and dataset.}\label{tab:prompts-responses-lengths}
\end{table}

For the user study, we removed
duplicate response-pairs (including pairs with identical responses,
see~\cref{subsec:appendix:user-study-identical-responses}), and were left
with 756 unique response-pairs, each of which was rated by five or more
participants. 

The prompts were on average 13.33 tokens long and the responses were
52.80 tokens long. We break down the 
lengths for
prompts drawn from the two datasets and responses generated by the three
models in~\cref{tab:prompts-responses-lengths}.

\subsection{Response-Pair Similarity}\label{subsec:results:sentence-pair}

We first compare user ratings of the similarity between LLM response-pairs
to semantic similarity metrics used in prior work. Semantic similarity
metrics are a core component of sampling-based consistency metrics. As
defined in~\cref{subsec:methods:machine-metrics}, we refer to such
comparison as \emph{response-pair similarities}.

\begin{table}
    \centering
    \footnotesize
    \begin{tabular}{llcccc}
        \toprule
        \textbf{LLM}                       & \textbf{Dataset}          & \textbf{\bert}                   & \textbf{\bleu}          & \textbf{\rouge}         & \textbf{\usescore}               \\
        \midrule
        \multirow{2}{*}{\textbf{\gemma}}   & \coqa                     & 0.82                             & 0.82                    & \textbf{0.83}           & 0.82                             \\
                                           & \cellcolor{gray!20}\lmsys & \cellcolor{gray!20}0.58          & \cellcolor{gray!20}0.63 & \cellcolor{gray!20}0.59 & \cellcolor{gray!20}\textbf{0.68} \\
        \midrule
        \multirow{2}{*}{\textbf{\llama}}   & \coqa                     & 0.84                             & 0.84                    & 0.89                    & \textbf{0.91}                    \\
                                           & \cellcolor{gray!20}\lmsys & \cellcolor{gray!20}0.60          & \cellcolor{gray!20}0.60 & \cellcolor{gray!20}0.64 & \cellcolor{gray!20}\textbf{0.70} \\
        \midrule
        \multirow{2}{*}{\textbf{\mistral}} & \coqa                     & \textbf{0.66}                    & 0.55                    & 0.65                    & 0.53                             \\
                                           & \cellcolor{gray!20}\lmsys & \cellcolor{gray!20}\textbf{0.57} & \cellcolor{gray!20}0.57 & \cellcolor{gray!20}0.52 & \cellcolor{gray!20}0.51          \\
        \midrule
        \textbf{All}                       & \textbf{Both}             & 0.71                             & 0.74                    & 0.73                    & \textbf{0.75}                    \\
        \bottomrule
    \end{tabular}
    \caption{At the response-pair level (not consistency), \usescore have the
        highest Spearman $\rho$ correlation coefficient with human ratings
        overall. Additionally, existing metrics better correlate with human
        evaluations for prompts from the \coqa dataset than \lmsys dataset.}
    \label{tab:response-pairs-metrics-spearman}
\end{table}

We evaluated the semantic similarities in a response-pair using
previously published methods and computed their correlation with human
ratings.
Each response-pair was rated by at least five participants. We
calculated Krippendorff's Alpha after removing the highest and lowest
score (see~\cref{subsec:methods:machine-metrics}) and found moderate agreements ($\alpha=0.72$) across
raters~\cite{krippendorff_computing_2011,marzi_kalpha_2024}.
Comparing each of the four
metrics for evaluating the similarity of response-pairs---\bert, \bleu, \rouge, and
\usescore~\cite{zhang_bertscore_2020,papineni_bleu_2001,lin_rouge_2004,cer_universal_2018}---to
participants' ratings, we found that no single metric correlated best with how
participants rated
response-pairs' similarities across all datasets and models. Notably,
\usescore had the highest correlation coefficient with response-pairs
produced by Llama, while \bert performed the best for Mistral.

Interestingly, we found the correlation between participants' ratings and the five
metrics from previous work to be higher for prompts from the \coqa dataset than
the \lmsys dataset. Participants found response-pairs created from prompts
from the \coqa dataset to be more similar to each other than response-pairs
created from \lmsys prompts, by one level in the 6-point scale. To the best of our
knowledge, \bert, \bleu, \rouge, and \usescore
have not been used to evaluate LLM consistency
with the more open-ended \lmsys dataset, which could have contributed to
their lower correlation with human ratings.

\begin{resultbox}
    \textbf{Result 1:} Among four existing metrics for evaluating the
      consistency of LLM responses based on similarity between pairs of
      sentences, \usescore best approximates human ratings
      in most cases. Additionally,
      existing metrics better approximate human ratings on \coqa than
      on \lmsys.
\end{resultbox}

\subsection{Response to Response-Set Consistency}
\label{subsec:results:response-consistency}
\begin{table*}
    \centering
    \small
    \begin{tabular}{lcccccccc|c}
        \toprule
                                 & \multicolumn{2}{c}{\textbf{DLR}} & \multicolumn{2}{c}{\textbf{Entropy}} & \multicolumn{2}{c}{\textbf{Prob}} & \multicolumn{2}{c|}{\textbf{LogProb}} & \multirow{2}{*}{\textbf{\usescore}}                                      \\
        Stats                    & Mean                             & Min.                                 & Mean                              & Max.                                  & Mean                                & Min. & Mean & Max. &               \\
        \midrule
        \textbf{Spearman $\rho$} & 0.37                             & 0.22                                 & 0.71                              & 0.75                                  & 0.69                                & 0.70 & 0.70 & 0.70 & \textbf{0.80} \\
        \textbf{MSE}             & 0.10                             & 0.05                                 & 0.41                              & 3.88                                  & 0.04                                & 0.14 & 0.33 & 0.67 & \textbf{0.02} \\
        \bottomrule
    \end{tabular}
    \caption{At response to response-set level, logit-based scores (showing 2
        best ones per type) did not perform as well as \usescore in
        approximating human ratings.\label{tab:prompt-logits}}
\end{table*}

In \cref{subsec:results:sentence-pair}, we evaluated methods for measuring the
similarity between pairs of responses. Consistency metrics, however, typically
involve computing a score for each response that represents the extent to which
that response is similar to or different from other responses to the same prompt.
Here, we compare how human-rating-based consistency metrics perform in
relation to those calculated with response-pair similarity metrics, as well
as one additional metric that requires sampling multiple responses,
\semanticentropy~\cite{kuhn_semantic_2023}. As defined
in~\cref{subsec:methods:machine-metrics}, we call this
\emph{response to response-set consistency}.

Motivated by reducing the cost of generating multiple responses to
a prompt for consistency estimation, we further explore whether logit-based
consistency-estimations methods (which have negligible cost compared to
response generation) can match human-based consistency scores.
For this investigation, we used 16 different logits based scores---mean,
minimum, maximum, and sum for each of DLR, Entropy, Probability, and
LogProbability (explained in~\cref{para:methods:logits})---and
found none of them individually approximates
the consistency of LLMs as well as the \usescore score (see~\cref{tab:prompt-logits}).

\begin{figure}
    \includegraphics[width=\columnwidth]{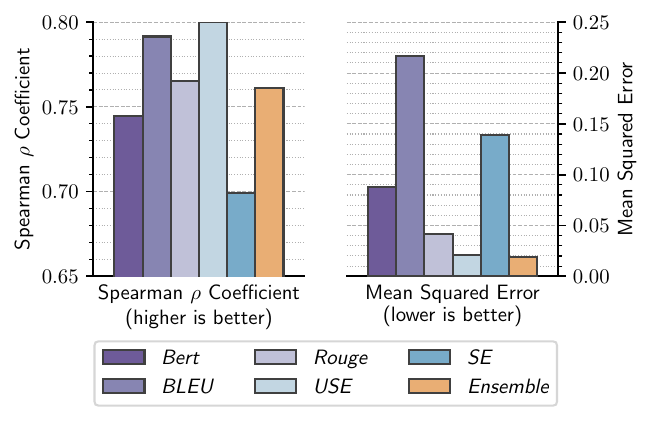}
    \caption{At response to response-set level, our ensemble of 16
    logit-based scores is as close of an approximation of human ratings as \usescore. 
    }
    \label{fig:response_metrics}
\end{figure}

As described in~\cref{para:methods:ensemble}, we used an ensemble of the 16
logit-based scores to attempt to approximate human ratings.
To determine which combination of the 16 logit-based scores
can create the best ensemble, 
we used \emph{Sequential Feature Selection}
(SFS)~\footnote{\url{https://scikit-learn.org/stable/modules/generated/sklearn.feature_selection.SequentialFeatureSelector.html}}.
We ran SFS 1,600 times, performing 100 repetitions at each selection size
(from using 1 logit-based score up to using all 16 scores).
Across the 1,600 runs, we found maximum
entropy to be most frequently selected (1,569 times) followed by sum of
LogProbability (1,384 times). We provide the full list
in~\cref{tab:features-selected-frequency}. Within each of the 1,600 runs,
we used 10-fold cross validation and evaluated the performance (i.e.,
Spearman $\rho$ and MSE) of our ensemble relative to human ratings. We
found that using all 16 logit-based scores resulted in the highest Spearman
correlation coefficient and lowest mean squared error when compared to
human ratings (see~\cref{fig:scores-selection}).
We found our ensemble method with 16 logit-based scores had a higher
correlation with human ratings than \bert, \bleu, \rouge, and \semanticentropy.
Our ensemble, when compared to the human ratings, performed as well as \usescore,
with a 0.002 better MSE~(\cref{fig:response_metrics}). As
in~\cref{subsec:results:sentence-pair}, we found that at the response to response-set
level, existing metrics correlate with human ratings
better on the \coqa dataset than on \lmsys (see~\cref{fig:responses-spearman}).

In addition to matching the performance of the best existing metrics, our
ensemble method benefits from reduced computational cost. With the same
GPU, it took \bert 2,649 seconds, \usescore 23 seconds, \rouge 14 seconds,
\semanticentropy 47 seconds, and our ensembling method 0.0044 seconds to
evaluate the consistency of 1,000 responses to 100 prompts. Training the
ensemble with 10-fold cross-validation on 90\% of the responses required an
average of 0.037 seconds across 100 runs using all 16 features.

\begin{resultbox}
    \textbf{Result 2:}
    Using an ensemble of 16 logit-based scores can produce a consistency
    estimate that approximates human ratings as accurately as the best
    existing metrics. In addition to matching \usescore, our method greatly
    reduces computational cost by not requiring the generation of multiple
    responses.
\end{resultbox}

\subsection{Prompt Level Consistency}
\label{subsec:res:consistency-by-prompt} 
As evidenced by the difference in how existing metrics perform on the two
different prompt datasets (\cref{tab:response-pairs-metrics-spearman}),
prompt choice is
a major contributor to consistency.

For example, responses to the factual question ``Where was the first modern
Olympic Game?'' were objectively more consistent than responses to the
open-ended question ``How can LLMs help users in their daily life?''
As such, we next investigate how to best measure consistency for a given prompt. Specifically,
we aggregate response-level consistency
from~\cref{subsec:results:response-consistency} to obtain a single
consistency score per prompt.

\begin{figure}
  \includegraphics[width=\columnwidth]{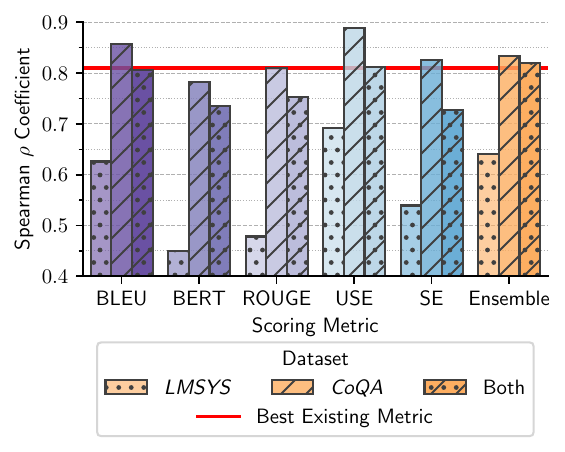}
  \caption{At per-prompt level, among existing metrics, our logit-based ensemble
    method and \usescore have the highest Spearman correlation coefficient
    with human evaluation of model-prompt consistency.}
  \label{fig:prompts-spearman-corr-metrics}
\end{figure}

\paragraph{Existing metrics} Using the definition from~\cref{subsec:methods:machine-metrics}, 
we calculate the consistency
score of each prompt using participants' ratings and metrics from previous
work. We found that \usescore-based consistency scores best correlate with
human-based consistency scores across all prompts
(\cref{fig:prompts-spearman-corr-metrics}). The consistency
scores on the \coqa dataset correlates better to human ratings than the consistency scores
on \lmsys. This is similar to the bias we observed with
pair-wise comparisons in~\cref{subsec:results:sentence-pair}.

\paragraph{Logit-based ensemble method} Our logit-based ensemble method,
described
in~\cref{para:methods:ensemble,subsec:results:response-consistency},
performs as well as \usescore, when aggregated to the prompt level, in
approximating human ratings of per-prompt consistency. While it did not
outperform \usescore, our method benefits from not needing to generate a
set of responses and compare responses within the set to determine LLMs'
responses consistency at the prompt level.

We further investigated the number of responses needed to train an ensembled
model with logit-based scores to predict the consistency of LLM responses
based on generating a single response.
However, due to the
relatively small number of prompts and responses in our sample, we were
not able to conclude how many responses are needed for such model to
accurately predict LLMs' response consistency at the prompt level. 

\begin{resultbox}
    \textbf{Result 3:} Our ensemble method using logit-based scores performs
    as well as existing metrics in approximating human ratings of LLM prompt-level
    consistency. This approach enables estimation of LLM consistency without the
    need to generate multiple responses to a prompt.
\end{resultbox}

\section{Conclusion and Discussion}\label{sec:discussion}

Using the human evaluation data, obtained 
via our user study, and subsequent experiments, 
we show that existing (automated) formulations of LLMs' consistency
are meaningfully different from our human-judgment
baseline. The best existing response to response-set 
consistency method achieves Spearman's $\rho=0.80$, indicating correlation 
but not representation. Further, prior work has suggested that correlations 
considered ``strong'' in less precise contexts~\cite{schober2018correlation} 
are not as meaningful for measuring 
success on NLP tasks~\cite{deutsch2022methods, DBLP:journals/corr/abs-2406-18403, shen2023large}.
This result holds regardless of whether we examine consistency 
at the per-response level (\cref{subsec:results:response-consistency}) or at the prompt level
(\cref{subsec:res:consistency-by-prompt}).  

Though sampling-based
methods come closest to the  
human baseline, we show that an ensemble of logit-based methods can approximate
this performance ($\rho_{ensemble}=0.82$ vs $\rho_{\text{\usescore}}=0.81$),
creating an opportunity to avoid the sampling overhead.

We also find (\cref{subsec:results:sentence-pair}) that the
difference in human judgement-based uncertainty vs.\ existing metrics for
LLM consistency approximation is greater for real-world prompt datasets
(\emph{LMSYS}) than for artificial ones (\emph{CoQA}). Determining why this is the
case is out of scope for our work, but we speculate that the research
community's focus 
on artificial testing datasets might be a contributing factor. 

\paragraph{Future directions}
The discrepancy between the human baseline and existing automated methods
of measuring LLM consistency 
raises important questions: How do imperfect consistency estimation 
methods affect downstream tasks? How would end-users be affected 
if shown consistency
metrics~\cite{kapoor_large_2024}? 
How are models affected when consistency 
measurements are part of the 
training cycle~\cite{liu2024can}? 
The answers are unclear and could be investigated by future work.

Finally, we urge researchers 
to consider the human evaluation
baseline in future research on how to measure LLM consistency, as well as in research that utilizes these metrics for downstream tasks.
We further recommend they (also) use real-world prompts for evaluation.

\section*{Limitations}\label{sec:limits} 
Since our investigation required
logit-based scores, which are often not accessible with 
black-box models, we used
open-weights models. Such usage limits the 
generalizability of our findings to
black-box models. Additionally, the prompts and 
responses in our sample are
relatively short (see~\cref{tab:prompts-responses-lengths}) 
despite using realistic prompts 
from users~\cite{zheng_lmsys-chat-1m_2023}; therefore,
our findings may not generalize to prompts and 
responses of all lengths. Lastly,
user studies are expensive to conduct.
While we recruited almost 3,000
users, we were only able to evaluate 100 prompts 
and 10 responses per prompt.
As we alluded to in~\cref{sec:discussion}, more data 
from users are needed to
establish a robust baseline for measuring LLM consistency.

\section*{Ethical Considerations}\label{sec:etchics}
Our study was approved by our institution's ethics review board. For the user
study, we first provided an informed consent form to participants explaining the
purpose of our survey, expected length, risks and benefits, as well as
compensation. Participants who gave consent to participate proceeded to the
survey. As explained in~\cref{subsec:methods:prompts-responses}, responses shown
to participants within the survey were filtered by the authors to remove 
potentially harmful (e.g., harassment, sexual, violence) content.

\section*{Distribution of Data and Artifact}
Code developed for this project, anonymized data collected from
participants, and analysis results are accessible at \url{https://doi.org/10.17605/OSF.IO/T9BF5}.

\section*{Acknowledgments}
This work was supported in part by the National Institute of Standards and
Technology (NIST) (\href{https://ror.org/05xpvk416}{ror.org/05xpvk416}) and
the Carnegie Mellon University
(\href{https://ror.org/05x2bcf33}{ror.org/05x2bcf33}) AI Measurement
Science and Engineering Center (AIMSEC); and by the PNC Center for
Financial Services Innovation at Carnegie Mellon University. We thank
Clement Fung for technical help with the ensemble predictor.

\bibliography{bibliography}

\appendix
\section{Appendix}\label{sec:appendix}

\subsection{Instructions for Survey Participants}\label{subsec:appendix:user-study-inst}
\paragraph{Risks} The primary risk is a breach of confidentiality since we use a
third-party (Qualtrics) to design our survey and collect survey responses.
Additionally, we utilize third-party vendors such as Prolific to recruit
participants, and Google Drive to store and process survey responses. This risk
is similar to what you encounter anytime you provide identifiable and private
information online. The risks and discomfort associated with participation in
this study are no greater than those ordinarily encountered in daily life or
other online activities. Participants might encounter boredom or fatigue.

\begin{figure*}[t]
    \includegraphics[width=\linewidth]{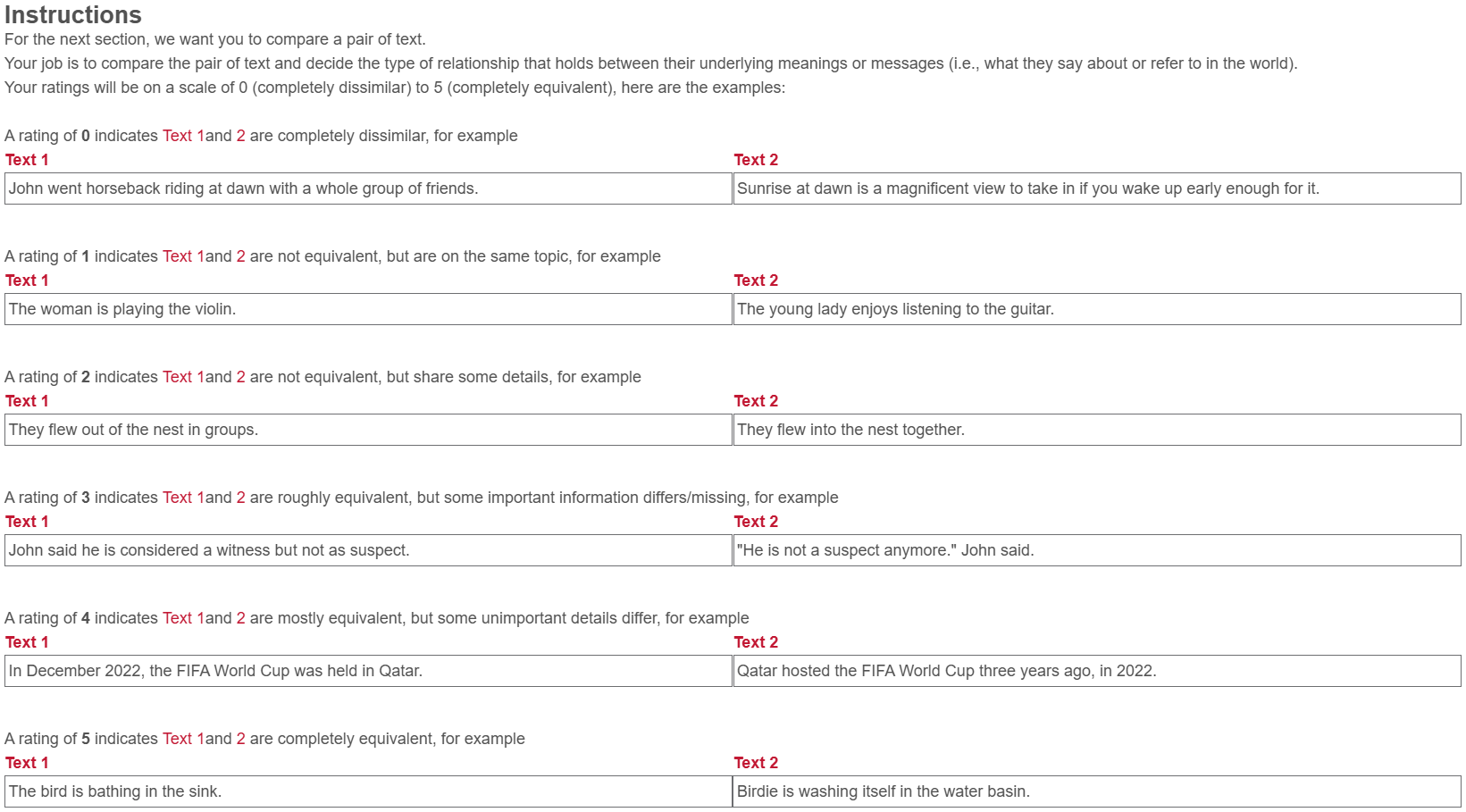}
    \caption{Instructions provided to participants for comparing pairs of sentences.}
    \label{fig:instructions}
\end{figure*}

\paragraph{Instructions} A screenshot of the instructions given to participants
is provided (\cref{fig:instructions}).

\subsection{Experiemnt asking users to rate identical responses}\label{subsec:appendix:user-study-identical-responses}
For each prompt, the 10 sampled responses described
in~\cref{subsec:methods:prompts-responses} are not all unique. We ran a 20
participant experiment on Profific with 5 pairs of unique responses for 5
different prompts. We found 17 participants gave a \emph{5 (completely
equivalent)} for the identical responses. Further, three participants each
gave a \emph{4}, \emph{3}, and \emph{0} to three different pairs of
identical responses. This result show participants are highly likely to
rate identical pairs of responses as a \emph{5 (completely equivalent)} so
we excluded all identical pairs of responses from further study.

\subsection{Supplemental Tables and Figures}

\begin{table}
    \centering
    \small
    \begin{tabular}{lr}
        \toprule
        \textbf{Logit-Based Score Name} & \textbf{Frequency} \\
        \midrule
        Max. Entropy & 1569 \\
        Sum. LogProb. & 1384 \\
        Mean Prob. & 1309 \\
        Sum. DLR & 1220 \\
        Min. DLR & 1123 \\
        Mean Entropy & 1044 \\
        Mean DLR & 984 \\
        Sum. Prob. & 856 \\
        Sum. Entropy & 691 \\
        Min. Entropy & 678 \\
        Max. DLR & 662 \\
        Min. Prob. & 555 \\
        Max. LogProb. & 533 \\
        Max. Prob. & 414 \\
        Mean LogProb. & 344 \\
        Min. LogProb. & 234 \\
        \bottomrule
    \end{tabular}
    \caption{Across 1,600 SFS runs (16 features $\times$ 100 repetitions),
    the most frequently selected features were Mac. Entropy, Sum. LogProb.,
    and Mean Prob.\label{tab:features-selected-frequency}}
\end{table}

\begin{table}[htbp]
    \centering
    \footnotesize
    \begin{tabular}{lrr}
        \toprule
        \textbf{Age} & \textbf{Num.} & \textbf{\%} \\
        18-24 & 290 & 9.74 \\
        25-34 & 917 & 30.81 \\
        35-44 & 708 & 23.79 \\
        45-54 & 556 & 18.68 \\
        55-64 & 324 & 10.89 \\
        65+ & 159 & 5.34 \\
        Prefer not to say & 22 & 0.74 \\
        \midrule
        \textbf{Gender} & \textbf{Num.} & \textbf{\%} \\
        Woman & 1538 & 51.68 \\
        Man & 1398 & 46.98 \\
        Other & 27 & 0.91 \\
        Prefer not to say & 13 & 0.44 \\
        \midrule
        \textbf{Education} & \textbf{Num.} & \textbf{\%} \\
        Associate's degree & 221 & 7.43 \\
        Bachelor's degree & 1225 & 41.16 \\
        Doctorate degree & 137 & 4.60 \\
        High school graduate & 298 & 10.01 \\
        Master's degree & 532 & 17.88 \\
        No high school degree & 11 & 0.37 \\
        Professional degree & 59 & 1.98 \\
        Some college credit, no degree & 420 & 14.11 \\
        Trade, technical, vocational training & 67 & 2.25 \\
        Other & 1 & 0.03 \\
        Prefer not to say & 5 & 0.17 \\
        \midrule
        \textbf{Income} & \textbf{Num.} & \textbf{\%} \\
        Under \$25K & 311 & 10.45 \\
        \$25K to \$50K & 582 & 19.56 \\
        \$50K to \$75K & 636 & 21.37 \\
        \$75K to \$100K & 470 & 15.79 \\
        \$100K or more & 915 & 30.75 \\        
        Prefer not to say & 62 & 2.08 \\
        \bottomrule
    \end{tabular}
    \caption{Demographics data of 2,976 participants.}\label{tab:demo}
\end{table}

Table~\ref{tab:demo} shows participants' demographics information.

\begin{figure}
    \includegraphics[width=\columnwidth]{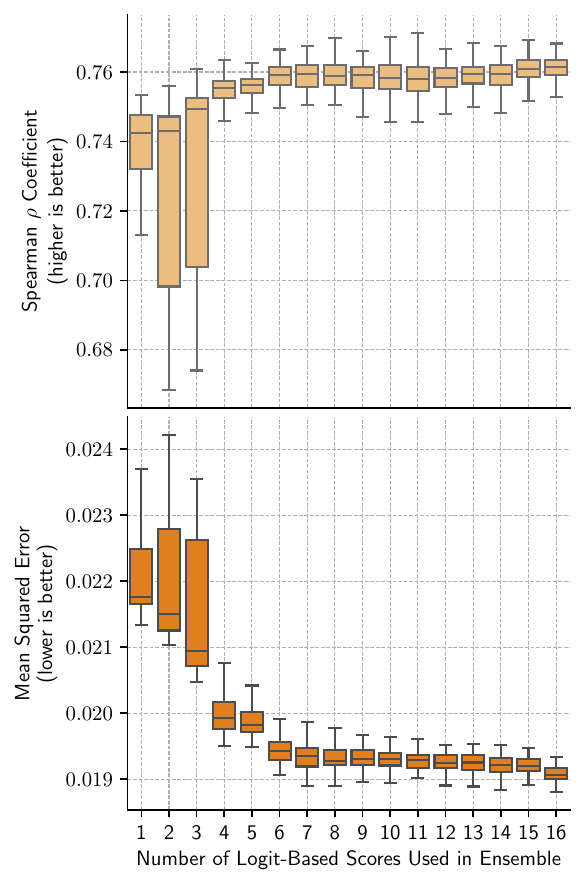}
    \caption{Running 100 10-fold cross validation shows using an ensemble of all
    16 logits-based scores yield the lowest Mean Squared Error and highest
    Spearman $\rho$ coefficient when compared to human ratings.}
    \label{fig:scores-selection}
\end{figure}

Figure~\ref{fig:scores-selection} shows our procedure of selecting all 16
logit-based scores in our ensemble method.

\begin{figure}
    \includegraphics[width=\columnwidth]{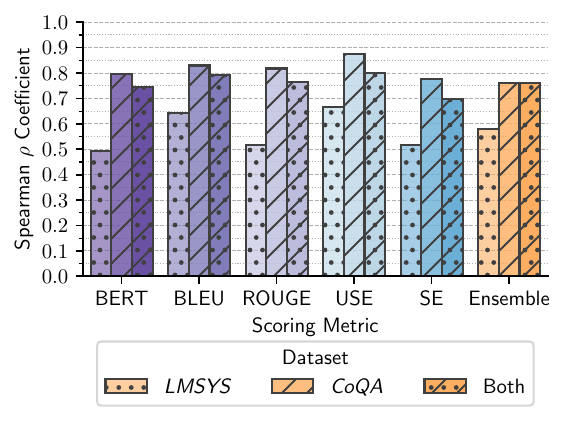}
    \caption{At response to response-set level, existing metrics and our
     ensemble method correlates better with human ratings on \coqa than \lmsys.}
    \label{fig:responses-spearman}
\end{figure}

Figure~\ref{fig:responses-spearman} shows the response to response-set level
comparison between existing metrics and our ensemble method across different
datasets.

\subsection{Example Prompts}
Table~\ref{tab:sample-prompts} shows prompts that had large discrepancies
between human rated consistency and consistency calculated with existing
metric (i.e., \usescore).
\onecolumn
\begin{table}[htbp]
    \centering
    \begin{tabularx}{\textwidth}{>{\raggedright\arraybackslash}p{0.4\textwidth}
    >{\raggedright\arraybackslash}X >{\raggedright\arraybackslash}X
    >{\centering\arraybackslash}p{0.15\textwidth} >{\centering\arraybackslash}p{0.15\textwidth}}
        \toprule
        \textbf{Prompt} & \textbf{Dataset} & \textbf{LLM Used} &
        \textbf{User Rated Consistency} & \textbf{\usescore Consistency} \\
        \midrule
        What has Williams become with this win? & CoQA & Gemma & 0.69 &
        0.42 \\
        Please resume the book "a libertarian walks into a bear" & LMSYS &
        Gemma & 0.50 & 0.28 \\
        Write to me something about Introversion and how I feel peace with
        small circle of people. & LMSYS & Llama & 0.77 & 0.51 \\
        Write a recipe for a unique dessert. & LMSYS & Mistral & 0.30 &
        0.63 \\
        \bottomrule
    \end{tabularx}
    \caption{Sample prompts with large differences between humans' ratings and the best surrogate metric (i.e., \usescore).}
    \label{tab:sample-prompts}
\end{table}

\end{document}